%% file: main.tex
\documentclass{article}

\PassOptionsToPackage{numbers, sort}{natbib}

\usepackage[preprint]{neurips_2026}

\usepackage[utf8]{inputenc} 
\usepackage[T1]{fontenc}    
\usepackage{hyperref}       
\usepackage{url}            
\usepackage{booktabs}       
\usepackage{amsfonts}       
\usepackage{nicefrac}       
\usepackage{microtype}      
\usepackage{xcolor}         

\usepackage{amsmath}
\usepackage{amssymb}
\usepackage{mathtools}
\usepackage{amsthm}
\usepackage{enumitem}
\usepackage{tabularray}
\UseTblrLibrary{booktabs}
\usepackage{subcaption}
\usepackage{adjustbox}
\usepackage{wrapfig}
\usepackage{algorithm}
\usepackage{algpseudocode}
\usepackage{afterpage}
\usepackage[normalem]{ulem}
\usepackage[capitalise]{cleveref}

\newcommand{\proposed}{\textsc{Sunta}}

\title{SUNTA: Hierarchical Video Prediction \\ with Surprise-based Chunking}

\author{%
  Tomoshi Iiyama \quad Masahiro Suzuki \quad Yutaka Matsuo \\
  The University of Tokyo \\
  \texttt{\{iiyama, masa, matsuo\}@weblab.t.u-tokyo.ac.jp} \\
}

\begin{document}

\maketitle

\begin{abstract}

\input{sections/0-abstract}

\end{abstract}


\input{sections/1-introduction}


\input{sections/2-method}

\input{sections/3-related-work}


\input{sections/4-experiments}

\input{sections/5-conclusion}

\bibliographystyle{plainnat}
\bibliography{references}


\newpage
\appendix

\input{sections/6-appendix}


\end{document}

%% file: sections/0-abstract.tex
Hierarchical state-space models (HSSMs) offer a promising approach to long-horizon prediction by segmenting sequences into temporal chunks.
However, their performance hinges on how chunk boundaries are determined.
While prior HSSMs typically rely on fixed-length chunking or similarity-based boundary detection, these methods often misalign with the intrinsic temporal structure of the data.
We argue that chunking should instead be driven by prediction errors, which more directly indicate when longer-range context becomes necessary. 
Nevertheless, integrating surprise-based chunking into HSSMs introduces critical challenges, including hierarchical collapse during end-to-end training and the absence of surprise signals during open-loop prediction.
To address these issues, we propose Surprise-based Nested Temporal Abstraction (\proposed{}), a method that employs a decoupled training strategy to preserve surprise signals and uses internal inconsistency as a top-down surprise metric to determine chunk boundaries within imagined rollouts.
Experiments on video prediction tasks in 2D and 3D environments demonstrate that \proposed{} outperforms baselines, uniquely maintaining accurate predictions over 250 timesteps, whereas all baselines degrade within the first 10 timesteps.

%% file: sections/1-introduction.tex
\section{Introduction} \label{sec:intro}

Predicting future environmental states is a cornerstone of intelligence~\citep{hawkins-intelligence, predictive-mind}, motivating research on world models that endow agents with internal predictive models~\citep{ha-world-model, lecun-path, FRISTON2021573} and substantially enhance planning and decision-making~\citep{dreamer, dreamerv2, dreamerv3, dreamer4, tdmpc, tdmpc2}.
A central challenge for such models is to support long-horizon prediction: since complex tasks often require foresight to anticipate distant consequences~\citep{tdmodels, pertsch2020gcp, director}, capturing long-term dynamics is essential for practical world models.

\begin{figure}[tbp]
  \centering
  \includegraphics[width=\textwidth]{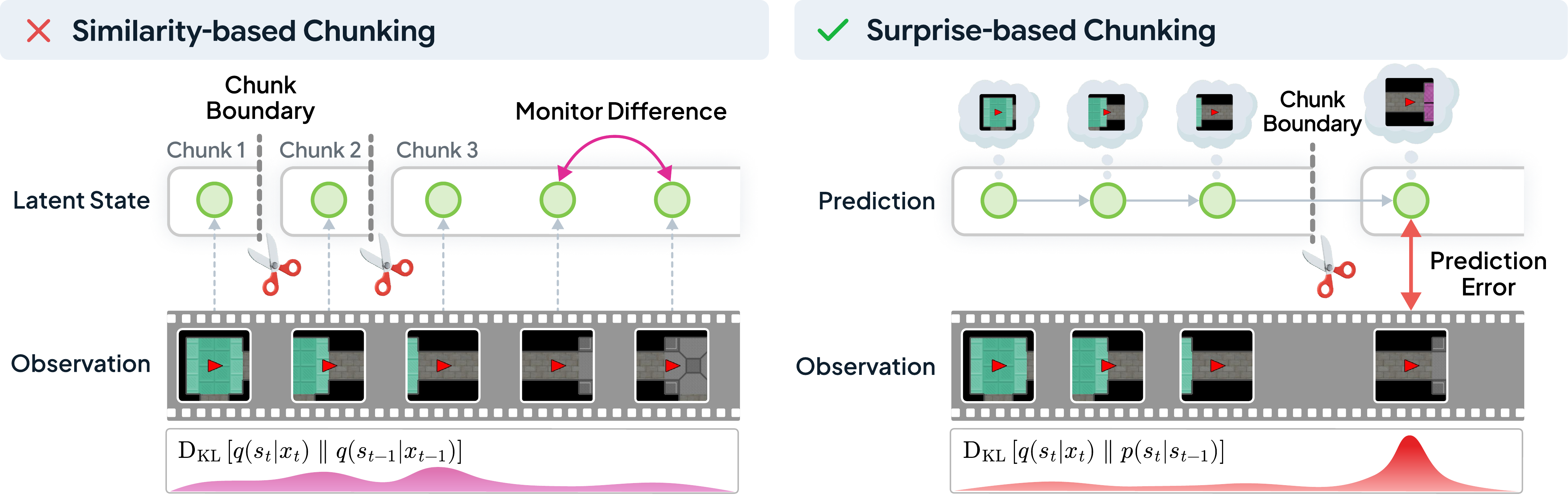}
  \caption{
  \textbf{Similarity-based chunking vs. surprise-based chunking}.
  Similarity-based chunking monitors observational changes and may over-segment superficial appearance changes or miss non-salient semantic transitions.
  Surprise-based chunking cuts at peaks of prediction error, highlighting shifts in the latent dynamics.
  }
  \label{fig:surprise-vs-similarity}
  \vspace{-20pt}
\end{figure}

A promising route toward long-horizon prediction is hierarchical state-space models (HSSMs)~\cite{vta, cwvae, mts3, thick, vpr, love}, which decompose sequences into temporal \emph{chunks} and learn dynamics across multiple timescales.
Crucially, performance depends on chunk boundaries: they should align with intrinsic temporal structure so that high-level dynamics are easier to model; misalignment forces the higher level to explain entangled transitions and obscures regularities even when the underlying dynamics are simple\footnote{For example, predicting a sentence becomes unnecessarily difficult when it is segmented into arbitrary substrings (\texttt{Iforg $\rightarrow$ otmyu $\rightarrow$ mbrel $\rightarrow$ laont $\rightarrow$ hetra $\rightarrow$ in}) rather than into syntactically meaningful units (\texttt{I $\rightarrow$ forgot $\rightarrow$ my $\rightarrow$ umbrella $\rightarrow$ on $\rightarrow$ the $\rightarrow$ train})}.

Nevertheless, existing HSSMs typically fix chunk lengths~\citep{mts3, cwvae} or constrain lengths toward a preset value~\citep{vta}.
Recent work, such as VPR~\citep{vpr}, instead infers boundaries from observational changes such as visual shifts (i.e., similarity-based chunking). However, such cues can fail in practice: important contextual shifts in the underlying dynamics may occur without visual salience, while superficial visual changes may appear without semantic transitions.

We argue that temporal abstraction should be driven by prediction errors, or \emph{surprise}, within the internal world model rather than visual changes, since surprise reflects shifts in the underlying dynamics rather than superficial appearance differences (see \cref{fig:surprise-vs-similarity}): chunks should begin and end where prediction error spikes, marking moments where higher-level abstraction should intervene.

The concept of surprise-based chunking dates back to early sequence modeling~\citep{schmidhuber1992surprise-chunker} and has been revisited for video understanding~\citep{aakur2019surprise-segmentation, mounir2023surprise-segmentation}. 
However, existing methods are mainly limited to \emph{recognition} settings.
While they segment data into chunks, they do not exploit those chunks for \emph{generation}, because naively introducing a surprise-based boundary criterion into HSSMs can lead to failure.
We identify two critical gaps that must be addressed to integrate surprise-based chunking into HSSMs:

\textbf{Gap 1: Hierarchical collapse under surprise minimization.}
In a naive implementation, low-level prediction errors are used to mark chunk boundaries, triggering transitions of high-level states.
However, the low level becomes \emph{less} surprised as the high-level model improves, because effective top-down conditioning makes the low-level dynamics easier to predict.
This erases the very surprise signals that defined the chunks, causing the hierarchy to cycle between emergence and collapse.

\textbf{Gap 2: Missing surprise during open-loop generation.}
Surprise signals rely on comparing predictions to external observations, limiting their use to the \emph{inference} or recognition process.
In open-loop rollouts of world models, we would like to leverage chunking to enable longer-horizon prediction; however, observations are inherently unavailable, so we cannot compute surprise-based chunk switches.

To address Gap~1, we treat each level as an independent module and optimize them \emph{sequentially}, so that improvements at the high level do not suppress the low-level surprise signal used for chunk discovery.
For Gap~2, we enable open-loop chunk detection by replacing bottom-up, observation-based surprise with a \emph{top-down} surprise signal derived from the high-level representation during generation.
We define this top-down surprise as the mismatch between the evolving low-level rollout and the current high-level context; as the mismatch grows, it triggers chunk termination and enables dynamic chunk detection.

Guided by these insights, we present an HSSM called \underline{\textbf{Su}}rprise-based \uline{\textbf{N}}ested \uline{\textbf{T}}emporal \uline{\textbf{A}}bstraction (\proposed{}).
To prevent hierarchical collapse, we adopt inter-level prediction: just as the low-level model predicts observations from the environment, the high-level model predicts low-level latents chunk by chunk, treating them as its observations.
By training each level sequentially, \proposed{} achieves stable optimization while maintaining surprise-driven hierarchical structures.
During open-loop generation, \proposed{} uses this top-down surprise to determine chunk boundaries without observations.

Finally, at test time (online inference or generation), the model must infer a high-level chunk representation from a \emph{partial} low-level chunk sequence, whereas during training it observes the \emph{entire} chunk.
To bridge this train--test mismatch, we introduce \emph{temporal pattern completion} regularization, training the high-level encoder to recover the chunk representation from incomplete low-level inputs and improving robustness to partial context.

We evaluate \proposed{} on three video datasets: a simple synthetic dataset with a moving ball, a 2D environment where an agent navigates across multiple rooms, and a 3D maze environment.
We evaluate the surprise-based criterion for accurately discovering meaningful temporal boundaries in videos and demonstrate the effectiveness of \proposed{} against 5 baseline models; in particular, \proposed{} is the only model that maintains accurate predictions for 250 timesteps, whereas all baselines degrade within the first 10 timesteps.

%% file: sections/2-method.tex
\begin{figure*}[tbp]
  \centering
  \includegraphics[width=\textwidth]{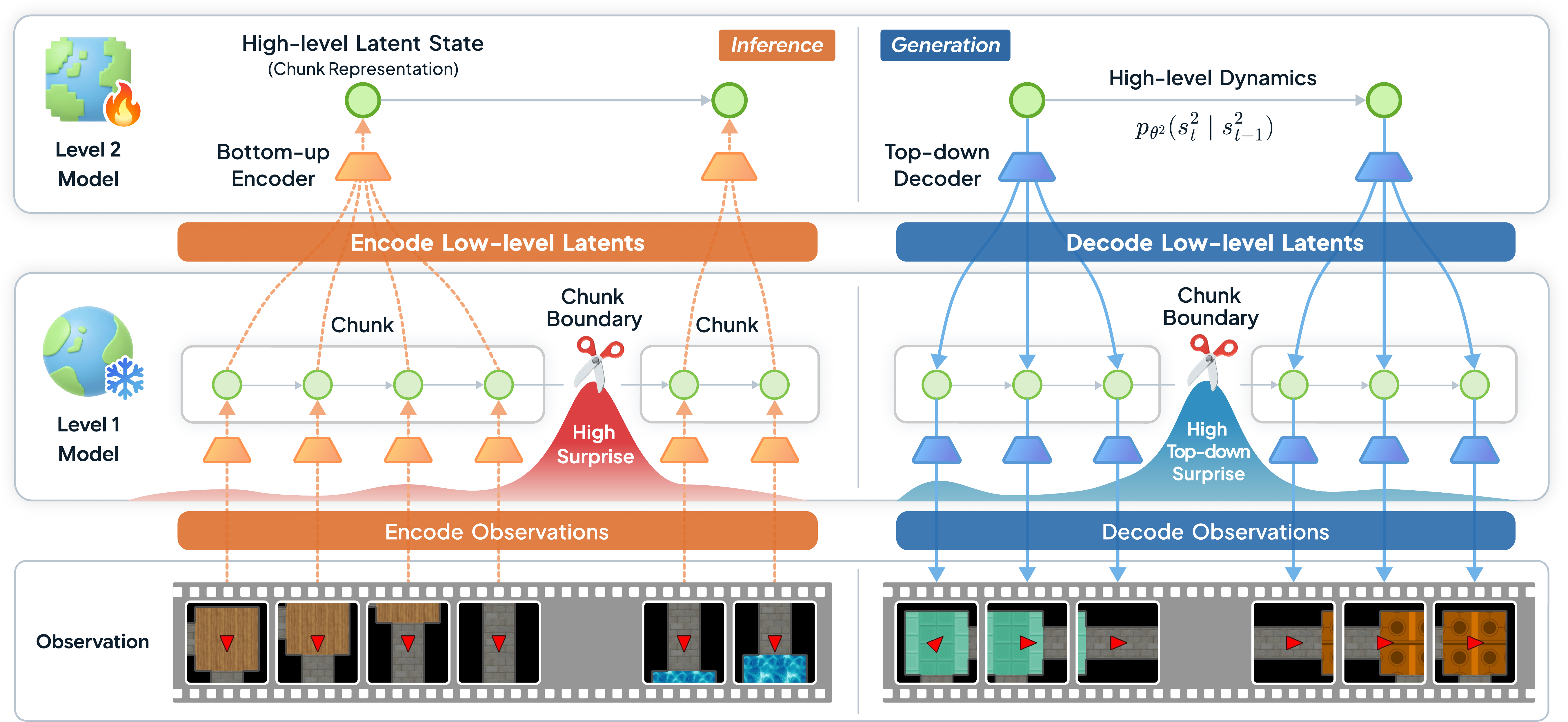}
  \caption{\textbf{Overview of \proposed{}}. During inference (left), observations are encoded into low-level states and segmented into chunks at points of high prediction error. A bottom-up encoder aggregates each chunk into a single latent state for high-level dynamics modeling. During generation (right), the high-level model produces high-level states that are decoded into low-level sequences. Chunk boundaries are detected via top-down surprise, which measures the mismatch between low-level rollouts and high-level context, enabling dynamic chunking without ground-truth observations.}
  \label{fig:overview}
  \vspace{-10pt}
\end{figure*}

\section{Preliminaries}

\paragraph{State-space models}
SSMs assume that observed sequences $x_{1:T}$ are generated by an underlying, hidden process of latent states $s_{1:T}$.
The model is defined by transition dynamics $p(s_t \mid s_{t-1})$, describing how the state evolves at each timestep $t$, and an observation model $p(x_t \mid s_t)$ that generates the observation conditioned on the latent state.

\paragraph{Hierarchical state-space models} 
HSSMs extend standard SSMs by introducing multi-level latent states that operate at multiple time scales.
They decompose a sequence of observations $x_{1:T}$ into temporal segments, referred to as chunks.
The low-level latent state $s^{(1)}_t$ operates at the frequency of raw observations (e.g., every video frame), while the high-level latent state $s^{(2)}_t$ remains constant within a chunk and is updated only at chunk boundaries. 
To formalize this temporal structure, we represent chunk boundaries using a binary variable $m_t \in \{0, 1\}$, indicating whether a new chunk starts at time step $t$~\citep{vta}.
If a boundary is not activated ($m_{t} = 0$), the high-level state is copied from the previous time step, persisting within the same chunk.
If a boundary is activated ($m_{t} = 1$), the high-level state transitions and a new chunk begins.

The critical difference among existing HSSMs lies in how the chunk boundary $m_t$ is determined. 
Existing approaches can be categorized into three main strategies:

\begin{itemize}[leftmargin=*,topsep=0pt,itemsep=0pt]
    \item \textbf{Fixed interval:} Fixes a chunk length $H$ as a hyperparameter, deterministically setting $m_t = 1$ if and only if $t \bmod H = 0$~\citep{cwvae, mts3}. 
    
    \item \textbf{Learning under constraints:} Treats $m_t$ as a latent variable inferred from observations via a variational posterior $q(m_{1:T} \mid x_{1:T})$~\citep{vta, love}.

    \item \textbf{Similarity-based:} Determines boundaries non-parametrically by setting $m_t = 1$ when the difference between successive latent representations exceeds a threshold~\citep{vpr}.
\end{itemize}

In this paper, we instead explore a surprise-based approach, where chunk boundaries are determined by the magnitude of prediction error in the low-level latents.

\section{Surprise-based Nested Temporal Abstraction} \label{sec:method}
We introduce surprise-based nested temporal abstraction (\proposed{}), a two-level HSSM that dynamically constructs its temporal hierarchy based on surprise signals.

\paragraph{Low-level model (level 1)}

We build our model upon the recurrent state space model (RSSM)~\citep{planet}, a backbone widely used in many HSSMs.
RSSM learns compact latent-state representations from high-dimensional observations using variational autoencoders, while modeling latent dynamics with recurrent neural networks (RNNs).
We consider an offline dataset consisting of sequences of observations \((x_1, \dots, x_T)\). 
We denote the low-level latent state at timestep \(t\) as \(s_t^{(1)}\), which is a stochastic variable parameterized as a diagonal Gaussian.
The low-level model of \proposed{}, parameterized by \(\theta_1\), consists of the following components:
\begin{align}
& \text{Posterior:} & &s^{(1)}_t \sim q_{\theta_1}(s^{(1)}_t \mid s^{(1)}_{<t}, x_t), \\
& \text{Prior:} & &\hat{s}^{(1)}_t \sim p_{\theta_1}(s^{(1)}_t \mid s^{(1)}_{<t}), \\
& \text{Decoder:} & &\hat{x}_t \sim p_{\theta_1}(x_t \mid s^{(1)}_t).
\end{align}
Here, the posterior is implemented with an autoregressive gated recurrent unit (GRU) that summarizes the latent history \(s^{(1)}_{<t}\), and is conditioned on an observation embedding obtained by a convolutional neural network (CNN) encoder applied to \(x_t\).
The prior uses a GRU to predict the next latent state from the same history, without access to \(x_t\).
The decoder reconstructs the observation from the stochastic latent state \(s^{(1)}_t\).

\paragraph{High-level model (level 2)}

Unlike typical HSSMs that treat all levels as a single neural network, \proposed{} employs a decoupled design where each level operates independently.
Analogous to the standard RSSM that models observations from the environment, the high-level model in \proposed{} treats the already learned low-level latents as \emph{observations}.
These low-level latents are segmented into chunks determined by the boundaries $m_t$, which are identified using the surprise criterion regarding the low-level predictions.
For each timestep $t$, we denote by $\tau(t)$ the starting timestep of the chunk to which $t$ belongs, so that the interval $[\tau(t), t]$ corresponds to the current chunk:
\begin{equation}
\tau(t) = \max\{\, t' \mid t' \leq t \;\wedge\; m_{t'} = 1 \,\}.
\end{equation}
Within a chunk, the high-level model learns a latent state $s_t^{(2)}$ that compresses the low-level sequence $s^{(1)}_{\tau(t):t}$ into a single representation.
The high-level model, parameterized by $\theta_2$, comprises the following components:
\begin{align}
& \text{Posterior:} & &s^{(2)}_t \sim q_{\theta_2}(s^{(2)}_t \mid s^{(2)}_{<t}, s^{(1)}_{\tau(t):t}), \\
& \text{Prior:} & &\hat{s}^{(2)}_t \sim p_{\theta_2}(s^{(2)}_t \mid s^{(2)}_{<t}), \\
& \text{Top-down Decoder:} & &\tilde{s}^{(1)}_{\tau(t):t} \sim p_{\theta_2}(s^{(1)}_{\tau(t):t} \mid s^{(2)}_t).
\end{align}
Here, $q_{\theta_2}$ is implemented with a GRU-based encoder that aggregates the low-level latent sequence within a chunk and outputs the parameters of the posterior over $s_t^{(2)}$.
The prior $p_{\theta_2}$ is implemented as a GRU that predicts the next high-level state from past high-level states without access to low-level latents.
The top-down decoder reconstructs the corresponding low-level latent sequence from $s_t^{(2)}$.

\paragraph{Surprise-based chunking}

We use the prediction error to determine chunk boundaries: the high-level model is activated only when the low-level model fails to predict the next input.
In probabilistic generative models, the prediction error can be quantified as Bayesian Surprise~\citep{surprise-taxonomy}, measured by the Kullback-Leibler (KL) divergence between the posterior and prior distributions.
We monitor the surprise measure in the low-level model at every time step, marking chunk boundaries whenever sharp increases in surprise are detected.
During \emph{inference}\footnote{Throughout this paper, we use the term ``inference'' in the classical generative-modeling sense of recognizing latent variables given observations (e.g., variational inference).} (recognition phase), we employ the following surprise criterion to identify chunk boundaries \(m_t\):
\begin{equation}
\mathrm{D_{KL}} \!\left[\,
q_{\theta_1}(s^{(1)}_t \mid s^{(1)}_{<t}, x_t)\ \big\|\ p_{\theta_1}(s^{(1)}_t \mid s^{(1)}_{<t})
\,\right] > \tau_{\text{inf}} .
\end{equation}
The threshold \(\tau_\text{inf}\) determines the sensitivity for detecting spikes.
While adaptive thresholds like moving averages could be used, in practice, we found that peak-detection algorithms that find local maxima in time series (e.g., SciPy's \texttt{find\_peaks}) work more robustly than simple moving-average thresholds.
See \cref{sec:sensitivity} for implementation details and sensitivity analysis.

\begin{figure*}[tp]
  \centering
  \begin{tblr}{
      colspec={l X[l]},
      width=\textwidth,
      rows={0mm}
  }
      Video & \adjustbox{valign=c}{\includegraphics[width=\linewidth]{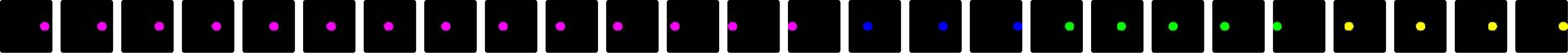}} \\
      Truth & \adjustbox{valign=c}{\includegraphics[width=\linewidth]{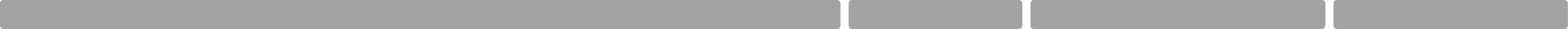}} \\
      \proposed{} & \adjustbox{valign=c}{\includegraphics[width=\linewidth]{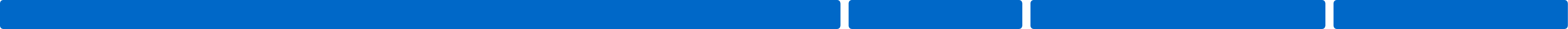}} \\
      VPR & \adjustbox{valign=c}{\includegraphics[width=\linewidth]{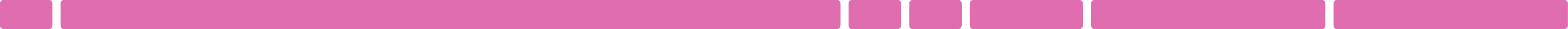}} \\
      VTA & \adjustbox{valign=c}{\includegraphics[width=\linewidth]{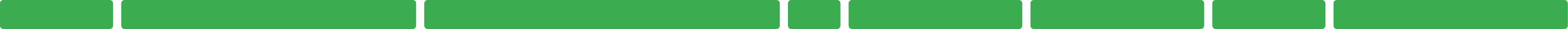}} \\
      LOVE & \adjustbox{valign=c}{\includegraphics[width=\linewidth]{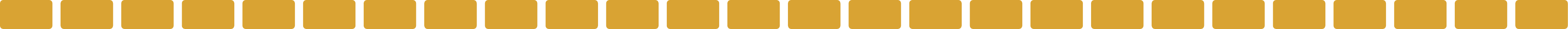}} \\
  \end{tblr}
\caption{\textbf{Hierarchical structures detected on the Bouncing Ball dataset}. Each colored block denotes a chunk identified by the corresponding model. Ground-truth boundaries occur when the ball bounces and changes color. \proposed{} precisely identifies these transitions, while VPR and VTA produce fragmented over-segmentation, and LOVE fails to detect meaningful boundaries altogether.}
\label{fig:BRL-boundary-results}
\end{figure*}

\paragraph{Level-wise training}

A primary obstacle in learning HSSMs lies in the circular dependencies between hierarchical levels that arise when attempting end-to-end learning involving both top-down and bottom-up processes within a Bayesian framework~\citep{friston-fast-slow}.
Consequently, many existing HSSMs bypass explicit bottom-up inter-level inference, instead inferring latent states at all levels conditioned solely on the raw environmental observations~\citep{cwvae, vta, love, vpr}.
\proposed{} adopts an alternative strategy: we train the hierarchy \emph{sequentially}, optimizing each level independently before moving to the next.
This decoupled optimization avoids circular dependencies and ensures that low-level latents and their boundaries are learned without being influenced by higher-level learning.

\paragraph{Open-loop chunk detection via top-down surprise}
During rollout, the high-level state provides chunk-level context while the top-down decoder generates the concrete low-level dynamics conditioned on it.
However, surprise signals normally require external observations, and are therefore unavailable in open-loop rollouts; as a result, we cannot compute bottom-up, observation-based surprise to decide when to switch chunks.
We address this by introducing a \emph{top-down surprise} signal that measures the mismatch between the evolving low-level rollout and the current high-level context.
Intuitively, as the rollout drifts beyond what the current high-level state represents, top-down surprise grows and signals that the chunk should terminate.
Specifically, at each step we first decode the current high-level state using the top-down decoder to generate the low-level latent state \(\tilde{s}^{(1)}_t\).
We then employ the following top-down surprise criterion to identify chunk boundaries:
\begin{equation}
\mathrm{D_{KL}} \left[
q_{\theta_2}\!\left(s^{(2)}_{t} \mid s^{(2)}_{<t}, \tilde{s}^{(1)}_{\tau(t):t} \right)
\ \parallel\
p_{\theta_2}\!\left(s^{(2)}_{t} \mid s^{(2)}_{<t} \right)
\right] > \tau_{\text{gen}},
\end{equation}
where \(\tau_{\text{gen}}\) controls the sensitivity for detecting chunk switches.

\paragraph{Temporal pattern completion (TPC)}
A key challenge in HSSMs is a train--test mismatch at higher levels.
During the training of a high-level model, the bottom-up encoder has access to the \emph{entire} lower-level chunk to infer the corresponding high-level latent state. 
However, during online inference or generation, the model must perform inference based on only a \emph{partial} chunk observed up to the current timestep.
This mismatch can yield inaccurate high-level representations from incomplete patterns, degrading the model's ability to reconstruct inputs or predict future transitions accurately.
To mitigate this, we introduce a regularizer that encourages the representation inferred from a partial chunk to match the more informative representation inferred from the complete sequence.
Concretely, we compare two posteriors over the high-level state: a \emph{complete} posterior inferred from the entire lower-level chunk and a \emph{partial} posterior inferred from a prefix of the same chunk.
We minimize the discrepancy between the \emph{complete} and \emph{partial} posteriors with a KL penalty, and add a contrastive separation term to prevent representational collapse by enforcing a margin from a \emph{negative} posterior.
Here, \(q_{\theta_2}^{\mathrm{c}}(s_t^{(2)} \mid \cdot)\), \(q_{\theta_2}^{\mathrm{p}}(s_t^{(2)} \mid \cdot)\), and \(q_{\theta_2}^{\mathrm{n}}(s_t^{(2)} \mid \cdot)\) denote the posteriors inferred from the \emph{complete} chunk, its \emph{partial} prefix, and an unrelated (\emph{negative}) chunk context (e.g., a different chunk/sequence), respectively.
The resulting temporal pattern completion loss is:
\begin{equation}
\begin{aligned}
\mathcal{L}_{\text{TPC}}(\theta_2)
&=\ \mathrm{D_{KL}} \!\Bigl[
q_{\theta_2}^{\mathrm{c}}(s_t^{(2)} \mid \cdot)
\parallel
q_{\theta_2}^{\mathrm{p}}(s_t^{(2)} \mid \cdot)
\Bigr] + \biggl[\gamma - \mathrm{D_{KL}} \!\Bigl[
q_{\theta_2}^{\mathrm{c}}(s_t^{(2)} \mid \cdot)
\parallel
q_{\theta_2}^{\mathrm{n}}(s_t^{(2)} \mid \cdot)
\Bigr]\biggr]_+,
\end{aligned}
\label{eq:tpc}
\end{equation}
where \([z]_+ = \max(0,z)\) and \(\gamma > 0\) is a margin hyperparameter.
This encourages high-level states that are robust to missing future context while remaining discriminative across different chunks.
We illustrate the train--test mismatch and our TPC procedure in \cref{sec:tpc-appendix}.

\paragraph{Model training}
As previously outlined, \proposed{} employs a level-wise training strategy where each level is trained independently, proceeding from the lowest level upward.
Initially, the low-level model (\(l=1\)) is trained on the raw observation data.
The training objective for the low-level model is equivalent to that of a standard RSSM, aiming to maximize the Evidence Lower Bound (ELBO):
\begin{equation}
\begin{aligned}
\mathcal{L}(\theta_1)
&= \mathbb{E}_{q_{\theta_1}(s^{(1)}_{\le T}\mid x_{\le T})} \Bigg[
\sum_{t=1}^{T} \log p_{\theta_1}(x_t \mid s^{(1)}_t)
- \sum_{t=1}^{T} \mathrm{D_{KL}}\!\left(
q_{\theta_1}(s^{(1)}_t \mid s^{(1)}_{<t}, x_t)\ \parallel\ p_{\theta_1}(s^{(1)}_t \mid s^{(1)}_{<t})
\right)
\Bigg],
\end{aligned}
\end{equation}
which maximizes reconstruction likelihood while regularizing the posterior toward the prior.
After training, we freeze \(\theta_1\) and infer low-level latent sequences \(s^{(1)}_{1:T}\) (and boundaries \(m_{1:T}\)).
We then train the high-level model by treating these inferred low-level latents as \emph{observations}.
For the high-level model, the objective consists of (i) a reconstruction term for low-level latents, (ii) a KL regularizer for the high-level dynamics, and (iii) the temporal pattern completion regularizer:
\begin{equation}
\begin{aligned}
&\mathcal{L}(\theta_2)
= \mathbb{E}_{q_{\theta_2}(s^{(2)}_{\le T} \mid s^{(1)}_{\le T}, m_{\le T})} \Bigg[
\sum_{t=1}^{T} \log p_{\theta_2}(s^{(1)}_{\tau(t):t} \mid s^{(2)}_t) \\
&- \sum_{t=1}^{T} \mathrm{D_{KL}}\!\left(
q_{\theta_2}(s^{(2)}_t \mid s^{(2)}_{<t}, s^{(1)}_{\tau(t):t})
\parallel
p_{\theta_2}(s^{(2)}_t \mid s^{(2)}_{<t})
\right)
- \mathcal{L}_{\text{TPC}}(\theta_2)
\Bigg],
\end{aligned}
\end{equation}
where \(\mathcal{L}_{\text{TPC}}(\theta_2)\) is defined in Eq.~(\ref{eq:tpc}) and includes both the complete--partial alignment and the margin-based contrastive separation.

%% file: sections/3-related-work.tex
\section{Related Work}

\paragraph{Temporal abstraction in predictive models}
Temporal abstraction is recognized as a fundamental cognitive ability in animal intelligence~\citep{hierarchical-behavior}, and realizing this capability within machine learning has been a long-standing research goal~\citep{schmidhuber1992surprise-chunker, tdmodels, sutton-smdp}.
In time-series modeling, HSSMs have been developed to leverage temporal abstraction to improve long-term predictions.
Early work often used fixed timescales, where state transitions occur at predetermined frequencies~\citep{hrnn, cwrnn, fsrnn, cwvae, mts3, hieros}, but performance depends on how well these intervals align with the (often variable) timescales of data.
Recent research, therefore, has shifted towards HSSMs with adaptive timescales, which learn or dynamically infer the boundaries of temporal segments from the data itself~\citep{mtrnn, hmrnn, mr-hdmm, stm, vta, vpr, love, thick}.
For example, VTA~\citep{vta} introduces latent boundary variables but requires predefined hyperparameters to regulate chunk statistics, while VPR~\citep{vpr} detects boundaries via similarity-based changes and may miss boundaries without salient perceptual shifts.
In contrast, our approach detects boundaries independently of these changes.
LOVE~\citep{love} improves upon VTA by incorporating additional objectives to minimize the code length of high-level latent states.
THICK~\citep{thick} learns adaptive timescales by enforcing \(L_0\) regularization on low-level latent states.

\paragraph{Chunking}
The cognitive process of organizing repeated information into smaller, meaningful units, known as chunking, is a fundamental mechanism of human learning and memory~\citep{miller-magical-number, chess-chunking}.
The principle of chunking has been widely adopted in compression algorithms~\citep{bpe}, sequence modeling~\citep{schmidhuber1992surprise-chunker}, representation learning~\citep{hcm}, tokenization for large language models~\citep{blt, megabyte, h-net}, and action prediction in robotics~\citep{q-chunking, rtc}.
However, despite these advances, no prior work has successfully integrated chunking mechanisms into predictive models for high-dimensional visual inputs.

\paragraph{Predictive coding}
Predictive coding posits that prediction errors in hierarchical generative models drive learning by continuously comparing predictions with sensory input~\citep{predictive-coding}.
Beyond serving as a learning signal, spikes in prediction error have been linked to the boundaries of meaningful events~\citep{event-perception}.
Early RNN work leveraged prediction failures in a lower-level network to trigger updates in a higher-level network, helping capture long-term dependencies~\citep{schmidhuber1992surprise-chunker}.
STM extends this idea to deep generative latent dynamics for visual inputs and improves planning, but it cannot generate all intermediate timesteps during imagination, limiting its use for standard video prediction~\citep{stm}.

\begin{table}[tbp]
    \small
    \caption{\textbf{Results on open-loop video prediction}. SSIM and MSE (mean $\pm$ std across 3 random seeds) of each model's predicted frames against ground-truth frames over 250 timesteps.}
    \vspace{6pt}
    \label{table:prediction-scores}
    \centering
    \begin{tblr}{
      colspec={X[l] c c c c c c},
      width=\textwidth,
      row{1} = {abovesep=1ex, belowsep=0.0ex},
      row{3} = {abovesep=1ex},
    }
      \toprule
      & \SetCell[c=2]{c}\textbf{Bouncing Ball} & & \SetCell[c=2]{c}\textbf{3D Maze} & & \SetCell[c=2]{c}\textbf{Serial Nine Rooms} \\
      & SSIM $\uparrow$ & MSE $\downarrow$ & SSIM $\uparrow$ & MSE $\downarrow$ & SSIM $\uparrow$ & MSE $\downarrow$ \\
      \midrule
      RSSM & 0.911 $\pm$ 0.014 & 1395 $\pm$ 25.6 & 0.635 $\pm$ 0.019 & 7750 $\pm$ 25.1 & 0.686 $\pm$ 0.004 & 2325 $\pm$ 72.1 \\
      VTA & 0.883 $\pm$ 0.004 & 1308 $\pm$ 16.2 & 0.471 $\pm$ 0.048 & 7355 $\pm$ 102 & 0.259 $\pm$ 0.029 & 5648 $\pm$ 371 \\
      LOVE & 0.887 $\pm$ 0.010 & 1285 $\pm$ 19.8 & 0.517 $\pm$ 0.027 & 7194 $\pm$ 104 & 0.206 $\pm$ 0.009 & 5895 $\pm$ 373 \\
      \proposed{} & \textbf{0.981 $\pm$ 0.004} & \textbf{210 $\pm$ 27.8} & \textbf{0.933 $\pm$ 0.011} & \textbf{580 $\pm$ 48.0} & \textbf{0.858 $\pm$ 0.025} & \textbf{507 $\pm$ 104} \\
      \bottomrule
    \end{tblr}
  \end{table}

  \begin{figure*}[bp]
    \small
    \centering
    \begin{tblr}{
        colspec={l X[l]},
        width=\textwidth,
        rows={5mm}
    }
        & \adjustbox{valign=c}{\includegraphics[width=\linewidth]{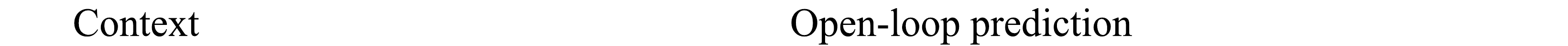}} \\
        Truth & \adjustbox{valign=c}{\includegraphics[width=\linewidth]{figures/BR-prediction-truth.pdf}} \\
        \proposed{} & \adjustbox{valign=c}{\includegraphics[width=\linewidth]{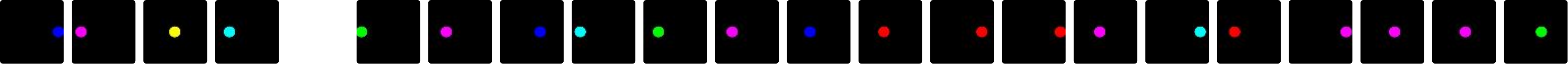}} \\
        RSSM & \adjustbox{valign=c}{\includegraphics[width=\linewidth]{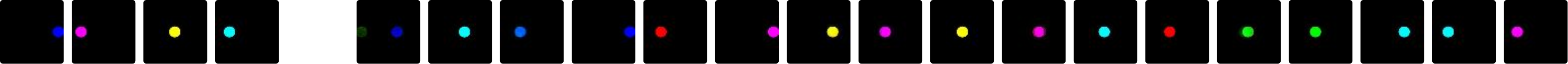}} \\
        VTA & \adjustbox{valign=c}{\includegraphics[width=\linewidth]{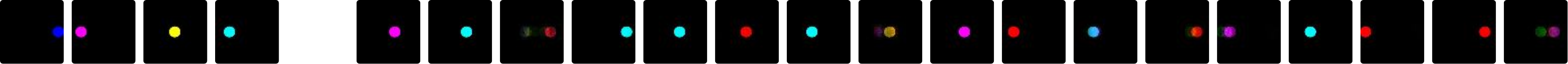}} \\
        LOVE & \adjustbox{valign=c}{\includegraphics[width=\linewidth]{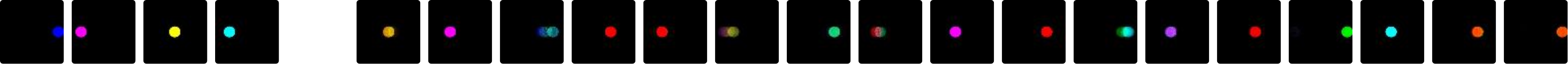}} \\
        & \adjustbox{valign=c}{\includegraphics[width=\linewidth]{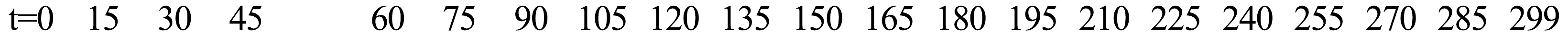}} \\
    \end{tblr}
  \caption{\textbf{Qualitative results of open-loop video prediction on the Bouncing Ball dataset}. The top row shows the ground-truth image frames and other rows show the image frames generated by each model up to 250 steps ahead, conditioning on the first 50 context frames.}
  \label{fig:BR-prediction-results}
  \end{figure*}

%% file: sections/4-experiments.tex
\section{Experiments} \label{sec:experiments}

We evaluate \proposed{} on two key questions: (1) Is surprise-based chunking effective for discovering temporal boundaries? (2) Does the resulting hierarchical structure improve long-term video prediction over prior methods?

\subsection{Datasets} \label{sec:datasets}
We evaluate \proposed{} on three video datasets with natural temporal hierarchies.
\textbf{Bouncing Ball} is a 2D toy dataset based on Bouncing Balls~\citep{vta, vpr}: a colored ball bounces off walls and changes color at each bounce, with the post-bounce color determined by the sixth previous color, introducing long-range dependencies.
\textbf{3D Maze} is the egocentric maze navigation dataset from VTA~\citep{vta}, extended with action-conditioned dynamics and wall colors that vary across episodes.
\textbf{Serial Nine Rooms}, inspired by~\citet{pertsch2020gcp} and built on Miniworld~\citep{MinigridMiniworld23}, places an agent traversing nine serially connected rooms with top-down egocentric observations; each room's texture is determined by the room visited three rooms earlier (${\sim}$120 timesteps), creating delayed dependencies.

\subsection{Baselines}

We compare \proposed{} against 5 video generative models.

\textbf{RSSM}~\citep{planet} is a non-hierarchical SSM, used to quantify the benefit of temporal abstraction.

\begin{wraptable}{r}[0pt]{0.43\textwidth}
  \vspace{-1pt}
      \small
      \caption{\textbf{Results on chunk boundary detection}. F1 scores (mean $\pm$ std across 3 random seeds) of the boundaries identified by each model during inference, evaluated against ground-truth boundary points.}
      \label{table:boundary-scores}
      \centering
      \begin{tblr}{
        colspec={X[l] c c},
        row{1} = {abovesep=4pt},
        row{3} = {abovesep=4pt},
      }
        \toprule
        & \textbf{Bouncing Ball}  & \textbf{3D Maze} \\
        & F1 Score $\uparrow$ & F1 Score $\uparrow$ \\
        \midrule
        VTA & 0.639 $\pm$ 0.082 & 0.577 $\pm$ 0.025 \\
        VPR & 0.467 $\pm$ 0.034 & 0.174 $\pm$ 0.039 \\
        LOVE & 0.276 $\pm$ 0.083 & 0.159 $\pm$ 0.008 \\
        \proposed{} & \textbf{0.984 $\pm$ 0.002} & \textbf{0.992 $\pm$ 0.003} \\
        \bottomrule
      \end{tblr}
      \vspace{-30pt}
\end{wraptable}

\textbf{VTA}~\citep{vta} is a 2-level HSSM that directly learns the chunk boundary.

\textbf{VPR}~\citep{vpr}, which learns a hierarchy via similarity-based chunking, cannot generate frames in an open-loop setting; we therefore use it only for boundary detection.

\textbf{LOVE}~\citep{love} augments VTA with an additional compression objective to improve chunk structure.

\textbf{\textsc{CW-Sunta}} 
To ablate our chunking mechanism, we introduce a fixed-timescale variant named Clockwork \proposed{}, analogous to CW-VAE~\citep{cwvae} and CW-VPR~\citep{vpr}.
It uses the same architecture and training as \proposed{}, but chunk boundaries are placed at predetermined fixed rates.
We set the chunk length to 10.

\begin{figure*}[tp]
  \centering
  \includegraphics[width=\textwidth]{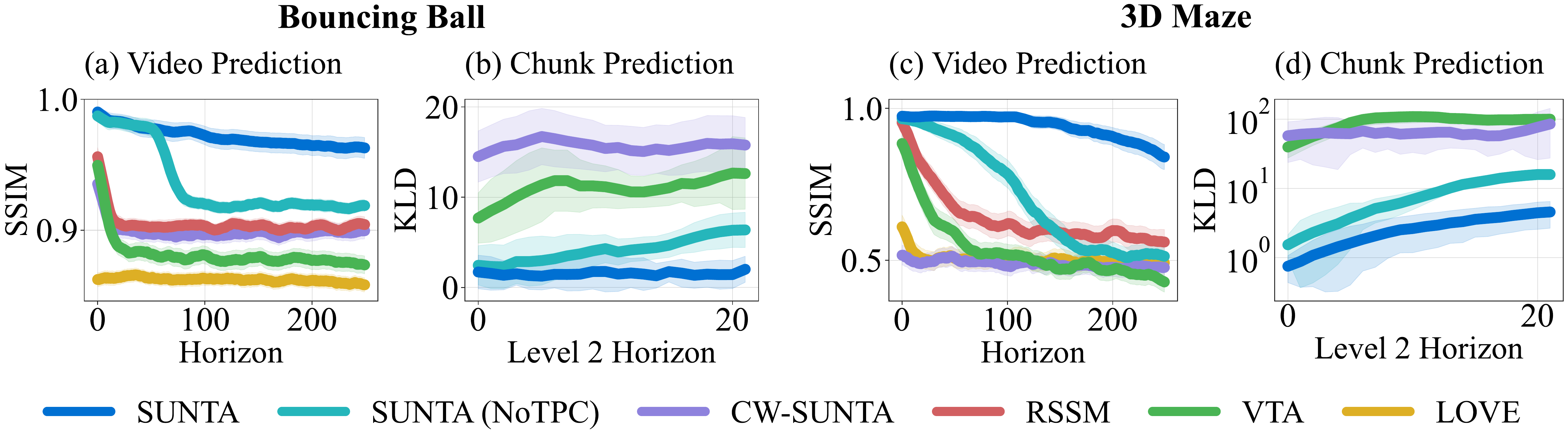}
  \caption{\textbf{Open-loop prediction accuracy over the rollout horizon.} We evaluate pixel-level quality via SSIM (higher is better $\uparrow$) and high-level latent-space prediction accuracy via KL divergence (lower is better $\downarrow$). All models are conditioned on the first 50 frames.}
  \label{fig:ssim-kl-over-horizon}
\end{figure*}

\subsection{Training and Evaluation Details}
All images are resized to \(64 \times 64\).
Models are trained with the ADOPT optimizer~\citep{adopt} on an NVIDIA H100.
We report F1 scores for boundary detection and the Structural Similarity Index (SSIM) and Mean Squared Error (MSE) for video prediction.
Quantitative results in \cref{table:boundary-scores,table:prediction-scores} are reported as mean $\pm$ standard deviation over 3 random seeds.

\subsection{Results and Discussion}

\paragraph{Results on Bouncing Ball}
We first evaluate the ability of \proposed{} to discover temporal structures.
For boundary detection, \cref{table:boundary-scores,fig:BRL-boundary-results} show that baselines struggle to identify ground-truth boundaries, whereas \proposed{} attains near-perfect boundary detection.
We next evaluate long-horizon prediction, conditioning on 50 context frames and generating the subsequent 250 frames (\cref{table:prediction-scores}).
RSSM fails to capture long-range dependencies, exposing the limitations of flat models.
VTA and LOVE, which failed to identify accurate boundaries (\cref{fig:BRL-boundary-results}), also perform poorly on this task.
None of the baselines recover the underlying color-transition rule.
In contrast, \proposed{} significantly outperforms all baselines.
\cref{fig:BR-prediction-results} further illustrates that all baselines degrade within 10 timesteps, while only \proposed{} maintains accurate predictions throughout the 250-frame horizon.
These results indicate that hierarchical structure significantly affects prediction performance, and that surprise-based chunking is the most reliable mechanism in this setting.

\begin{figure*}[t]
  \small
  \centering
  \begin{tblr}{
      colspec={l X[l]},
      width=\textwidth,
      rows={5mm}
  }
      & \adjustbox{valign=c}{\includegraphics[width=\linewidth]{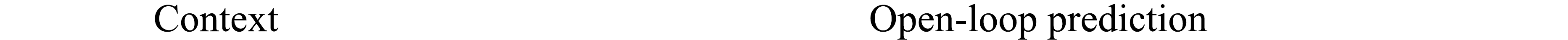}} \\
      Truth & \adjustbox{valign=c}{\includegraphics[width=\linewidth]{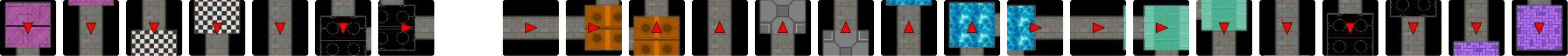}} \\
      \proposed{} & \adjustbox{valign=c}{\includegraphics[width=\linewidth]{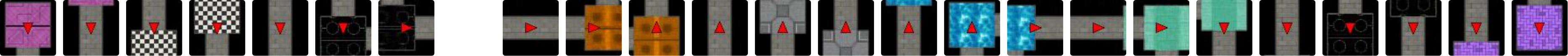}} \\
      RSSM & \adjustbox{valign=c}{\includegraphics[width=\linewidth]{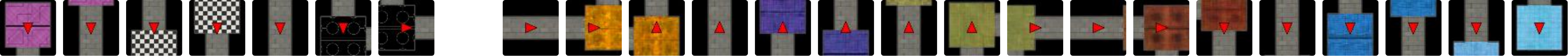}} \\
      VTA & \adjustbox{valign=c}{\includegraphics[width=\linewidth]{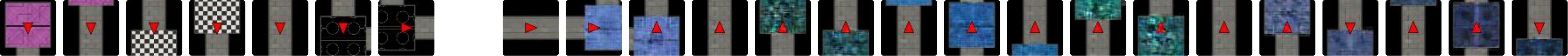}} \\
      LOVE & \adjustbox{valign=c}{\includegraphics[width=\linewidth]{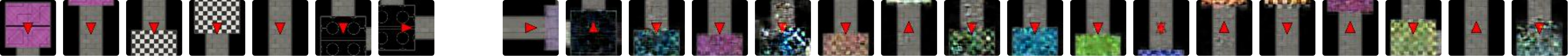}} \\
      & \adjustbox{valign=c}{\includegraphics[width=\linewidth]{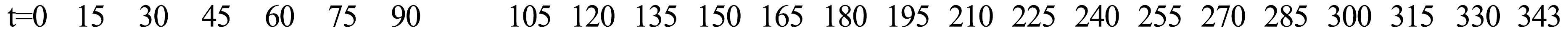}} \\
  \end{tblr}
\caption{\textbf{Qualitative results of open-loop video prediction on the Serial Nine Rooms dataset}. The top row shows the ground-truth image frames and other rows show the image frames generated by each model up to 244 steps ahead, conditioning on the first 100 frames.}
\label{fig:S9R-prediction-results}
\end{figure*}

\paragraph{Results on Serial Nine Rooms}
We further challenge the models on the Serial Nine Rooms dataset, which features more complex textures and longer-range dependencies.
\cref{table:prediction-scores} shows that the gap between \proposed{} and the baselines widens further on this dataset: while RSSM degrades substantially and the hierarchical baselines VTA and LOVE collapse altogether, \proposed{} maintains a clear margin over all of them.
As illustrated in \cref{fig:S9R-prediction-results}, \proposed{} generates coherent room-to-room transitions and remains visually consistent over the full horizon, whereas baselines quickly drift and produce mismatched textures, reflecting their difficulty in maintaining the delayed dependency.

\paragraph{Results on 3D Maze}
Finally, on 3D Maze, \proposed{} again dominates both boundary detection (\cref{table:boundary-scores}) and long-horizon prediction (\cref{table:prediction-scores}).
Notably, both hierarchical baselines (VTA, LOVE) underperform the non-hierarchical RSSM across all datasets.
This suggests that imposing an arbitrary or misaligned hierarchy is not merely ineffective but actively detrimental.
\proposed{} avoids this failure mode, successfully uncovering meaningful abstractions even under 3D perspective changes; qualitative open-loop predictions on this dataset are shown in \cref{fig:3MSM-prediction-results} (\cref{sec:3d-maze-prediction}).

\paragraph{Effect of temporal pattern completion.}
\cref{fig:ssim-kl-over-horizon} (\proposed{} (NoTPC)) shows that the gap from removing the TPC objective widens with the rollout horizon.
We attribute this to the high-level encoder overfitting to complete chunks, yielding unstable representations from partial observations and causing incorrect latent transitions that compound over long horizons.

\paragraph{Fixing chunk length.}
The fixed-timescale variant \textsc{CW-Sunta} (i.e., \proposed{} without surprise-based chunking) matches the flat RSSM in long-horizon prediction (\cref{table:ablation-independent}, \cref{fig:ssim-kl-over-horizon}).
This indicates that imposing a hierarchy with rigid chunk lengths delivers no gain over a flat SSM, and that the added complexity of a hierarchical mechanism rarely pays for itself unless guided by effective boundaries.
The substantial leap achieved by \proposed{}, by contrast, demonstrates that the advantage of hierarchical modeling lies not in the architecture itself but in aligning abstraction with the environment's intrinsic temporal structure, yielding temporal abstractions that simplify the learning of high-level dynamics.

%% file: sections/5-conclusion.tex
\section{Conclusion} \label{sec:conclusion}
We introduced surprise-based nested temporal abstraction (\proposed{}), which brings surprise-driven chunking to hierarchical video prediction via a decoupled hierarchy and top-down surprise signals. 
Experiments show that \proposed{} substantially improves long-horizon prediction over prior HSSM baselines, highlighting the importance of aligning hierarchical abstraction with intrinsic temporal structure. 
These results advance world models for settings that demand accurate long-term foresight.

\paragraph{Limitations}
\proposed{} has several limitations that we hope future work will address.
First, the method relies on the presence of meaningful surprise spikes in the low-level model: in environments with uniformly unpredictable dynamics, the surprise signal could lack discriminative peaks and fail to guide effective chunking.
Second, our experiments focus on simulated visual environments, and extending \proposed{} to real-world, higher-resolution video and multimodal observations is left to future work.
Third, while we evaluate up to two levels of hierarchy and find that additional levels are unwarranted on our current benchmarks (\cref{sec:hierarchy-depth}), \proposed{}'s design naturally extends to deeper hierarchies, and verifying its gains in environments with multi-scale long-range dependencies remains an open question.

Finally, incorporating action generation and decision-making into our temporal abstraction framework represents an exciting avenue for hierarchical planning and control.

%% file: sections/6-appendix.tex
\section{Hyperparameters}
\label{sec:hyperparameters}

\cref{table:hyperparameters} summarizes the hyperparameters used in our main experiments.
Unless otherwise stated, the same values are used across all datasets.

\begin{table}[h]
    \caption{\textbf{Hyperparameters used in the experiments.}}
    \vspace{4pt}
    \label{table:hyperparameters}
    \centering
    \begin{tblr}{
      colspec={X[l] c},
      width=\textwidth,
      row{1} = {abovesep=1ex, belowsep=0.0ex},
    }
      \toprule
      Hyperparameter & Value \\
      \midrule
      Optimizer & ADOPT~\citep{adopt} \\
      Learning rate & $3 \times 10^{-4}$ \\
      Batch size & 64 \\
      Training steps (per level) & 250{,}000 \\
      Gradient clipping & 10.0 \\
      \midrule
      Image resolution & $64 \times 64$ \\
      Recurrent (deterministic) state dim & 512 \\
      Stochastic state dim ($s^{(1)}_t$, $s^{(2)}_t$) & 32 \\
      \midrule
      Peak-detection prominence & 0.1 \\
      TPC margin $\gamma$ & 50 \\
      \midrule
      Loss weight: reconstruction (NLL)                          & 1.0 \\
      Loss weight: state KL (level-1 posterior--prior)           & 1.0 \\
      Loss weight: state reconstruction KL (level-2 to level-1)  & 1.0 \\
      Loss weight: TPC alignment (complete--partial KL)          & 1.0 \\
      Loss weight: TPC contrastive separation (margin term)      & 1.0 \\
      \bottomrule
    \end{tblr}
\end{table}

\section{Architecture Details}
\label{sec:architecture}

\paragraph{Vision encoder and decoder.}
We adopt a DreamerV2-style convolutional encoder--decoder for all models.
The encoder takes a $3 \times 64 \times 64$ RGB frame and applies four \texttt{Conv2d} blocks with kernel size $4$ and strides $(2, 2, 2, 1)$, each followed by \texttt{BatchNorm2d} and \texttt{ELU} activation, producing a $C$-dimensional embedding (channel width $C = 64$).
The decoder mirrors this with four \texttt{ConvTranspose2d} blocks with kernel size $4$ and strides $(1, 2, 2, 2)$ and \texttt{BatchNorm2d}+\texttt{ELU}.

\paragraph{RSSM internals.}
Each level is implemented as an RSSM~\citep{planet} with the following components:
\begin{itemize}[leftmargin=*,topsep=0pt,itemsep=0pt]
  \item Recurrent backbone: \texttt{GRUCell}.
  \item Latent distribution: diagonal Gaussian; the standard deviation is parameterized through a \texttt{Sigmoid} activation.
  \item Activation: \texttt{ELU} throughout the MLP heads.
\end{itemize}
The level-1 decoder takes the stochastic latent $s^{(1)}_t$ as input.
The level-2 top-down decoder takes the concatenation of the deterministic and stochastic components of $s^{(2)}_t$.

\section{Pseudocode}
\label{sec:pseudocode}

\proposed{} is trained in two stages following the decoupled level-wise scheme described in \cref{sec:method}.
\cref{alg:training} summarizes the procedure.

\begin{algorithm}[h]
\caption{\proposed{}}
\label{alg:training}
\begin{algorithmic}[1]
\State \textbf{Stage 1: train the low-level model.}
\State Initialize $\theta_1$.
\For{step $= 1, \dots, 250{,}000$}
  \State Sample a batch of observation sequences $(x_{1:T})$.
  \State Compute the level-1 ELBO and update $\theta_1$.
\EndFor
\State Freeze $\theta_1$.
\Statex
\State \textbf{Stage 2: train the high-level model.}
\State Initialize $\theta_2$.
\For{step $= 1, \dots, 250{,}000$}
  \State Sample a batch and infer $s^{(1)}_{1:T}$ via the frozen $q_{\theta_1}$.
  \State Compute the low-level surprise sequence and detect boundaries $m_{1:T}$ via \texttt{find\_peaks}.
  \State Sample a partial-chunk prefix and a negative-chunk context for the TPC regularizer.
  \State Compute the level-2 ELBO with $\mathcal{L}_{\text{TPC}}$ and update $\theta_2$.
\EndFor
\Statex
\State \textbf{Open-loop generation.}
\State Given a context segment, infer $s^{(2)}_t$ for the most recent chunk.
\For{rollout step $t' > t$}
  \State Decode top-down to obtain $\hat{s}^{(1)}_{t'}$ and roll the low-level prior forward.
  \State Compute the top-down surprise and detect chunk boundaries $\tilde{m}_{t'}$ via \texttt{find\_peaks}.
  \If{a boundary is detected}
    \State Transition the high-level state via the level-2 prior.
  \EndIf
\EndFor
\end{algorithmic}
\end{algorithm}

\section{Boundary Detection}
\label{sec:boundary}
\label{sec:sensitivity}

\paragraph{Surprise computation.}
We compute the inference-time surprise as the reverse KL divergence between the level-1 posterior and prior at each timestep (\cref{sec:method}).
We use the reverse formulation $\mathrm{D_{KL}}(q \,\|\, p)$ rather than $\mathrm{D_{KL}}(p \,\|\, q)$ because it more reliably detects shifts in the posterior toward unexpected outcomes.

\paragraph{Peak detection.}
We use SciPy's \texttt{find\_peaks} on the per-sequence surprise signal with the \texttt{prominence} parameter.
Surprise signals are first normalized by their running maximum to remove scale variation across sequences.
At the end of each sequence we append a sentinel value (the running mean of the signal) so that genuine surprise spikes at the final timestep are not missed by \texttt{find\_peaks}.

\paragraph{Hyperparameter sensitivity and choice of \texttt{prominence}.}
\proposed{} introduces a single peak-detection hyperparameter, the \texttt{prominence} parameter, used both during training to identify boundaries from low-level surprise (the threshold $\tau_\text{inf}$) and during open-loop generation to detect chunk switches from top-down surprise (the threshold $\tau_\text{gen}$).
\cref{table:sensitivity} reports the F1 of detected boundaries and the SSIM of open-loop video prediction across a wide range of \texttt{prominence} values on Bouncing Ball and 3D Maze.

We choose \texttt{prominence} $= 0.1$ for all main experiments because it sits near the center of the stable range $[0.05, 0.3]$ and transfers without retuning to Serial Nine Rooms, where ground-truth boundaries are unavailable for direct tuning.
Outside this range, we observe the failure modes consistent with their respective limits: very small values (e.g., $0.001$) trigger spurious peaks and over-segment the sequence, while very large values (e.g., $1.0$) miss genuine peaks and under-segment.
Both behaviors degrade downstream prediction quality.

\begin{table}[h]
    \caption{\textbf{Sensitivity to peak-detection prominence.} F1 score of detected boundaries during training (left) and SSIM of open-loop video prediction (right) on Bouncing Ball and 3D Maze, varying the \texttt{prominence} parameter of SciPy's \texttt{find\_peaks}. Performance is stable across a wide range; we use $0.1$ in all main experiments.}
    \vspace{4pt}
    \label{table:sensitivity}
    \centering
    \begin{tblr}{
      colspec={c c c c c},
      width=\textwidth,
      row{1} = {abovesep=1ex, belowsep=0.0ex},
    }
      \toprule
      & \SetCell[c=2]{c}\textbf{F1 score $\uparrow$} & & \SetCell[c=2]{c}\textbf{SSIM $\uparrow$} & \\
      \texttt{prominence} & Bouncing Ball & 3D Maze & Bouncing Ball & 3D Maze \\
      \midrule
      0.001 & 0.712 & 0.834 & 0.917 & 0.706 \\
      0.01  & 0.989 & 0.989 & 0.955 & 0.818 \\
      0.05  & 0.994 & \textbf{0.996} & \textbf{0.986} & 0.940 \\
      0.1   & \textbf{0.995} & \textbf{0.996} & \textbf{0.986} & \textbf{0.942} \\
      0.3   & \textbf{0.995} & \textbf{0.996} & 0.984 & 0.927 \\
      0.5   & 0.978 & 0.982 & 0.935 & 0.693 \\
      1.0   & 0.706 & 0.853 & 0.903 & 0.491 \\
      \bottomrule
    \end{tblr}
\end{table}

\section{Baseline Implementation}
\label{sec:baselines}

All baselines use the same vision encoder, decoder, recurrent backbone, and hidden dimensions as \proposed{} unless otherwise stated.

\paragraph{RSSM~\citep{planet}.}
A non-hierarchical baseline using only the level-1 components of \proposed{}.
To provide a stronger non-hierarchical baseline, we increase RSSM's capacity by setting the deterministic-state dimension to $2048$ and the stochastic-state dimension to $64$.

\paragraph{VTA~\citep{vta}.}
A two-level HSSM with learned binary boundary variables.
We use the authors' official implementation, with deterministic-state size $512$ and stochastic-state size $32$ at both levels.
We set the maximum number of subsequences \(N_\text{max}\) to $20$ and maximum subsequence length \(h_\text{max}\) to $15$ .

\paragraph{LOVE~\citep{love}.}
Built on top of VTA with an additional compression objective.
We use the authors' official implementation, with a discrete latent codebook of size $30$ and a coding-length coefficient of $0.1$.

\paragraph{VPR~\citep{vpr}.}
We re-implemented VPR following the original paper, using similarity-based chunking.
Because VPR cannot generate frames in an open-loop setting, we report it only for boundary detection.

\paragraph{CW-\proposed{}.}
A fixed-timescale ablation of \proposed{}.
The architecture and training procedure are identical to \proposed{}, but chunk boundaries are placed at fixed intervals with chunk length $10$.

\section{Compute Resources}
\label{sec:compute}

All experiments were run on a single NVIDIA H100 GPU.
Training each level for $250{,}000$ steps takes approximately 48 hours; full \proposed{} training (2-level) therefore takes approximately 96 hours per seed.

\section{Parameter Count}
\label{sec:complexity}

We compare \proposed{}'s parameter count against the baselines (\cref{table:complexity}).
\proposed{} achieves its gains with fewer trainable parameters than all baselines, indicating that its improvements arise from the surprise-driven chunking mechanism rather than from additional model capacity.

This efficiency stems from \proposed{}'s decoupled, inter-level inference (\cref{sec:method}).
Existing HSSMs infer latent states at all levels directly from the raw observations, so every level must process high-dimensional inputs.
In contrast, \proposed{}'s high-level model treats the already-inferred low-level latents as its observations and never processes the raw inputs, allowing it to achieve accurate long-horizon prediction with substantially less model capacity.

\begin{table}[h]
    \caption{\textbf{Parameter count comparison.} Trainable parameters at a fixed hidden dimension.}
    \vspace{4pt}
    \label{table:complexity}
    \centering
    \begin{tblr}{
      colspec={X[l] c},
      width=\textwidth,
      row{1} = {abovesep=1ex, belowsep=0.0ex},
    }
      \toprule
      Model & \# Parameters $\downarrow$ \\
      \midrule
      RSSM (flat)            & 30.74M \\
      VTA (2-level)          & 44.26M \\
      LOVE (2-level)         & 44.21M \\
      \proposed{} (2-level)  & \textbf{15.40M} \\
      \bottomrule
    \end{tblr}
\end{table}

\section{Temporal Pattern Completion}
\label{sec:tpc-appendix}

\begin{figure*}[h]
  \centering
  \begin{minipage}[t]{0.44\textwidth}
    \centering
    \includegraphics[width=.95\textwidth]{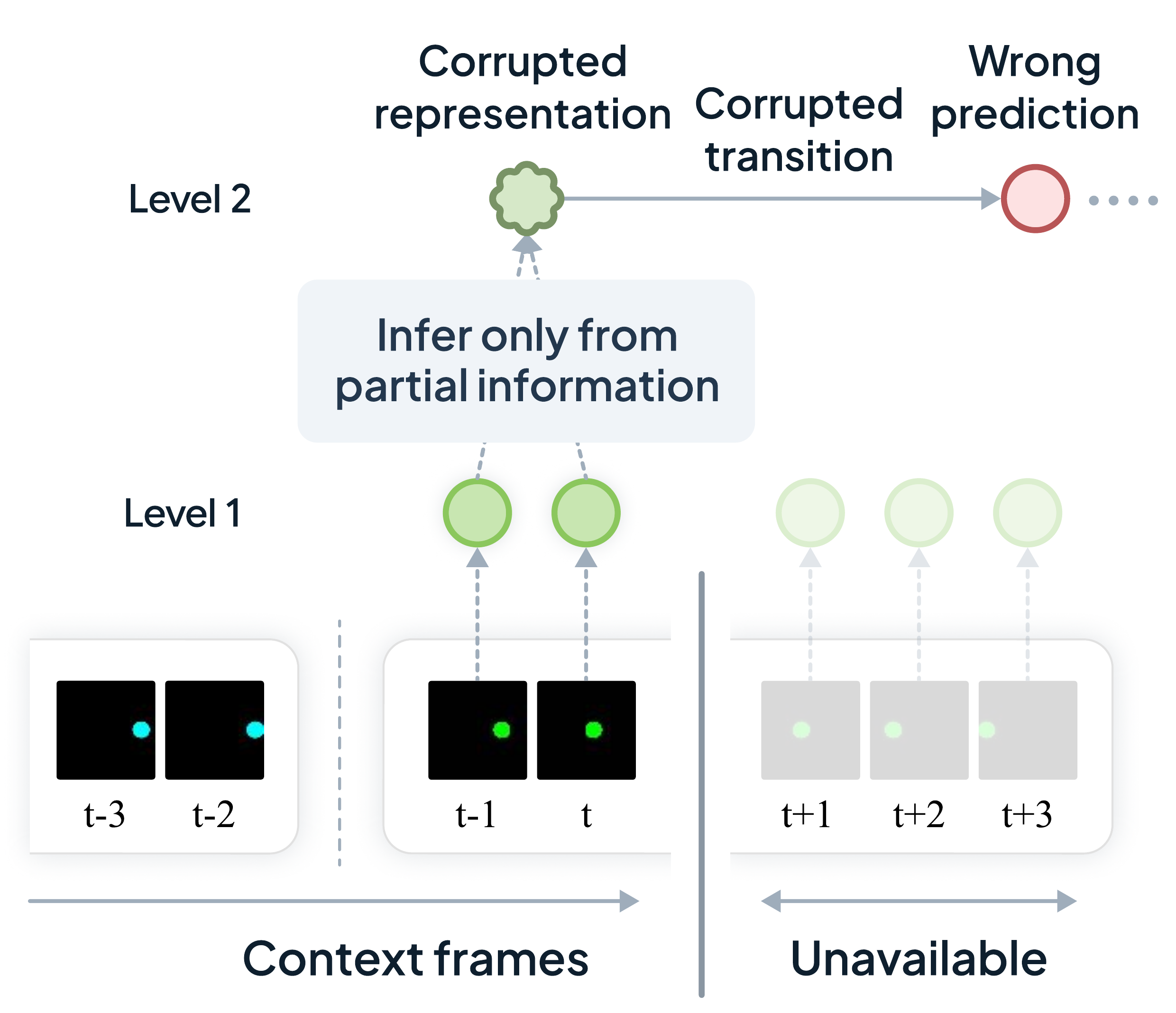}
    \subcaption{}
    \label{fig:TPC-problem}
  \end{minipage} \hfill
  \begin{minipage}[t]{0.55\textwidth}
    \centering
    \includegraphics[width=\textwidth]{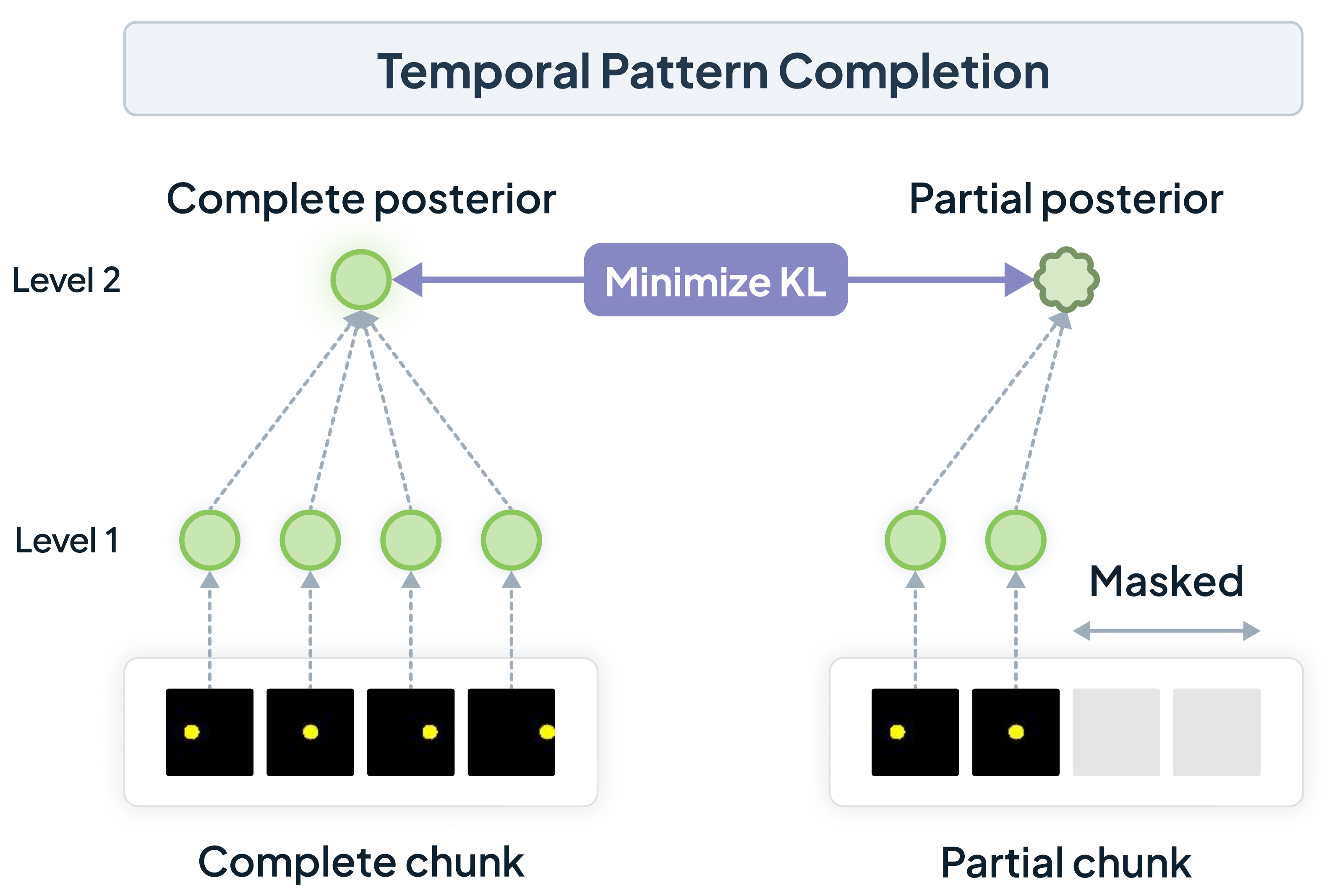}
    \subcaption{}
    \label{fig:TPC}
  \end{minipage}
  \caption{(a) \textbf{Incomplete chunks at the end of context.} During online inference, the context (observations available up to the current timestep) often ends with a truncated chunk. Since the high-level model is trained on complete chunks, this train--test mismatch can lead to flawed high-level representations and degrade subsequent prediction during the rollout. (b) \textbf{Temporal pattern completion (TPC) regularization.} To mitigate the train--test mismatch caused by partial chunks, we introduce TPC regularization for the high-level model. During training, we simulate a partial chunk by masking the final timesteps of a chunk. We then enforce representational consistency by minimizing the KL divergence between the high-level state inferred from the partial chunk and the state inferred from the complete chunk.}
  \label{fig:tpc}
\end{figure*}

\cref{fig:tpc} illustrates the train--test mismatch addressed by temporal pattern completion (TPC) and our regularization scheme (\cref{sec:method}).
At training time, the high-level encoder $q_{\theta_2}$ aggregates the entire low-level latent sequence within each chunk (\cref{fig:TPC-problem}, top) to produce the chunk representation $s_t^{(2)}$.
At test time, however, observations are revealed online: the most recent chunk in the context is typically truncated, and during open-loop generation the high-level state must be inferred from only a partial low-level rollout (\cref{fig:TPC-problem}, bottom).
Without explicit regularization, this discrepancy yields representations that drift away from those seen at training and propagate errors over the open-loop horizon.

To simulate this condition during training, we draw a prefix length uniformly within each chunk and feed the prefix through the high-level encoder to obtain the partial posterior $q^{\mathrm{p}}_{\theta_2}$ in addition to the complete posterior $q^{\mathrm{c}}_{\theta_2}$ (\cref{fig:TPC}).
The KL penalty in \cref{eq:tpc} aligns the partial posterior with the complete one, while the contrastive margin term against a negative posterior $q^{\mathrm{n}}_{\theta_2}$ drawn from an unrelated chunk prevents collapse to a chunk-invariant representation.

\section{Ablation Details}
\label{sec:ablation-details}

We conduct two complementary ablations on the Bouncing Ball dataset, isolating \proposed{}'s three core components: surprise-based chunking, decoupled level-wise training, and temporal pattern completion (TPC).

\paragraph{Independent ablation.}
\cref{table:ablation-independent} reports prediction SSIM when each component is removed individually from the full \proposed{} model.
Removing surprise-based chunking (yielding the fixed-timescale variant \textsc{CW-Sunta}) or TPC each causes a notable drop, while removing decoupled level-wise training is the most damaging single change.
\begin{table}[h]
    \caption{\textbf{Independent ablation on Bouncing Ball}. Open-loop video prediction SSIM when each core component is removed individually from the full \proposed{} model.}
    \vspace{4pt}
    \label{table:ablation-independent}
    \centering
    \begin{tblr}{
      colspec={X[l] c},
      width=\textwidth,
      row{1} = {abovesep=1ex, belowsep=0.0ex},
    }
      \toprule
      & \textbf{Bouncing Ball}  \\
      & SSIM $\uparrow$  \\
      \midrule
      \proposed{} (full) & \textbf{0.986} \\
      \quad w/o surprise-based chunking (i.e., \textsc{CW-Sunta}) & 0.918 \\
      \quad w/o temporal pattern completion (i.e., NoTPC) & 0.936 \\
      \quad w/o decoupled level-wise training & 0.880 \\
      \bottomrule
    \end{tblr}
\end{table}

\paragraph{Incremental ablation.}
\cref{table:ablation-incremental} starts from a non-hierarchical RSSM and successively adds each component.
Naively introducing a two-level hierarchy---whether with fixed chunks or with surprise-based chunking---under end-to-end training degrades performance below the flat RSSM, consistent with our hypothesis of hierarchical collapse.
Switching to decoupled level-wise training reverses this trend, and adding TPC yields a further substantial gain.
This pattern shows that surprise-based chunking and a two-level hierarchy on their own are insufficient: preventing hierarchical collapse via decoupled training is essential, and TPC further closes the partial-chunk train--test gap.

\begin{table}[h]
    \caption{\textbf{Incremental ablation on Bouncing Ball}. Open-loop video prediction SSIM as components are added on top of a non-hierarchical RSSM baseline.}
    \vspace{4pt}
    \label{table:ablation-incremental}
    \centering
    \begin{tblr}{
      colspec={X[l] c},
      width=\textwidth,
      row{1} = {abovesep=1ex, belowsep=0.0ex},
    }
      \toprule
      & \textbf{Bouncing Ball}  \\
      & SSIM $\uparrow$ \\
      \midrule
      RSSM (flat) & 0.904 \\
      $+\ $ End-to-end two-level RSSM (fixed chunk length) & 0.889 \\
      $+\ $ Surprise-based chunking & 0.883 \\
      $+\ $ Decoupled level-wise training & 0.936 \\
      $+\ $ Temporal pattern completion (= \proposed{}) & \textbf{0.986} \\
      \bottomrule
    \end{tblr}
\end{table}

\begin{figure*}[h]
    \centering
    \begin{tblr}{
        colspec={l X[l]},
        width=\textwidth,
        rows={0mm}
    }
        Video & \adjustbox{valign=c}{\includegraphics[width=\linewidth]{figures/BRL-boundary-frames.pdf}} \\
        Truth & \adjustbox{valign=c}{\includegraphics[width=\linewidth]{figures/BRL-boundary-truth.pdf}} \\
        \proposed{} (decoupled) & \adjustbox{valign=c}{\includegraphics[width=\linewidth]{figures/BRL-boundary-sunta.pdf}} \\
        \proposed{} (end-to-end) & \adjustbox{valign=c}{\includegraphics[width=\linewidth]{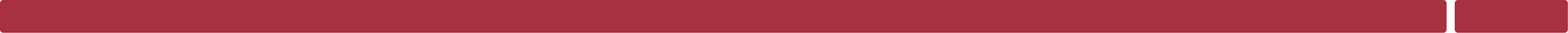}} \\
    \end{tblr}
  \caption{\textbf{Effect of decoupled training on chunk discovery (Bouncing Ball)}. End-to-end training collapses into a degenerate segmentation, whereas decoupled level-wise training preserves the surprise signal and recovers the ground-truth boundaries (where the ball bounces and changes color).}
  \label{fig:BRL-boundary-e2e}
\end{figure*}

\paragraph{Effect of decoupled training on chunk discovery.}
To diagnose hierarchical collapse, we visualize the chunks discovered by \proposed{} when trained end-to-end versus with the proposed decoupled procedure (\cref{fig:BRL-boundary-e2e}).
End-to-end training collapses to a degenerate segmentation that ignores the underlying color transitions, whereas decoupled training preserves the surprise signal and recovers the ground-truth boundaries.
This is consistent with the corresponding quantitative gain in \cref{table:ablation-incremental}.

\section{Additional Results}

\subsection{Generality across Backbones}\label{sec:backbones}
While we use GRU-based variational inference to match existing HSSM baselines, the fundamental concept of using prediction errors to form temporal chunks and learning chunk-level transitions with a separate model could be implemented with alternative backbone architectures.
To probe this, we replace the GRU in both levels with xLSTM~\citep{xlstm}, a more recent recurrent architecture.
\cref{table:backbone-comparison} shows that the gains of \proposed{} are preserved and slightly improved under this replacement, suggesting that surprise-based chunking transfers across recurrent backbones.
Extending \proposed{} to non-recurrent backbones such as Transformers is an interesting direction for future work.

\begin{table}[h]
    \caption{\textbf{Generality across recurrent backbones}. Open-loop video prediction on Bouncing Ball when replacing the GRU in both levels with xLSTM~\citep{xlstm}.}
    \vspace{4pt}
    \label{table:backbone-comparison}
    \centering
    \begin{tblr}{
      colspec={X[l] c c},
      width=\textwidth,
      row{1} = {abovesep=1ex, belowsep=0.0ex},
    }
      \toprule
      & SSIM $\uparrow$ & MSE $\downarrow$ \\
      \midrule
      \proposed{} w/ GRU   & 0.986 & 185 \\
      \proposed{} w/ xLSTM & \textbf{0.992} & \textbf{103} \\
      \bottomrule
    \end{tblr}
\end{table}

\subsection{Video Prediction on 3D Maze}
\label{sec:3d-maze-prediction}
\begin{figure*}[h]
    \centering
    \begin{tblr}{
        colspec={l X[l]},
        width=\textwidth,
        rows={5mm}
    }
        & \adjustbox{valign=c}{\includegraphics[width=\linewidth]{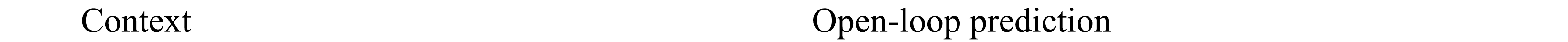}} \\
        Truth & \adjustbox{valign=c}{\includegraphics[width=\linewidth]{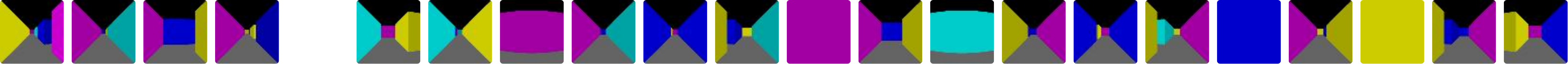}} \\
        \proposed{} & \adjustbox{valign=c}{\includegraphics[width=\linewidth]{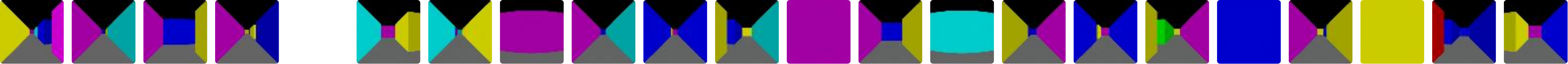}} \\
        RSSM & \adjustbox{valign=c}{\includegraphics[width=\linewidth]{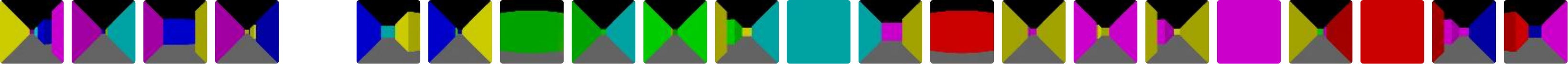}} \\
        VTA & \adjustbox{valign=c}{\includegraphics[width=\linewidth]{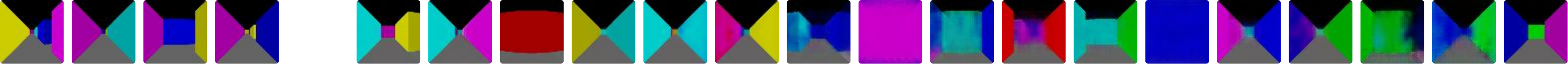}} \\
        LOVE & \adjustbox{valign=c}{\includegraphics[width=\linewidth]{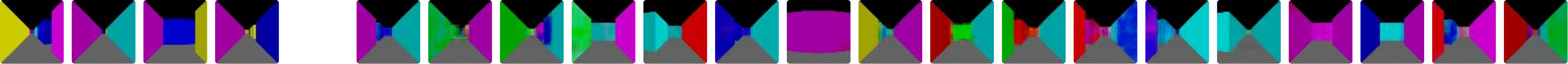}} \\
        & \adjustbox{valign=c}{\includegraphics[width=\linewidth]{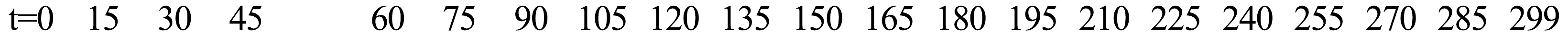}} \\
    \end{tblr}
  \caption{\textbf{Qualitative results of open-loop video prediction on the 3D Maze dataset}. The top row shows the ground-truth image frames and other rows show the image frames generated by each model up to 250 steps ahead, conditioning on the first 50 context frames.}
  \label{fig:3MSM-prediction-results}
\end{figure*}

\subsection{Prediction accuracy within high-level latent space}
To verify that our chunking mechanism simplifies the learning of high-level dynamics, we evaluate prediction accuracy solely within the high-level latent space. 
We measure the KL divergence between the ground-truth posteriors and the predicted priors obtained via recursive ``jumpy'' rollouts, thereby isolating the accuracy of chunk predictions from the image decoding quality or boundary detection errors during generation.
As illustrated in \cref{fig:ssim-kl-over-horizon}, baseline models exhibit consistently large KL divergence. 
Recalling their failure to identify correct boundaries (\cref{table:boundary-scores}), these results suggest that inappropriate chunking forces the high level to model incoherent transitions, making the dynamics difficult to capture even when the underlying temporal patterns are simple. 
Conversely, \proposed{} maintains low KL divergence throughout the rollout. 
This confirms that surprise-based chunking successfully simplifies the explanation of chunk transitions, yielding a tractable abstraction that directly translates to superior long-term prediction.

\section{Determining the Number of Hierarchical Levels}
\label{sec:hierarchy-depth}

\begin{figure*}[h]
    \centering
    \includegraphics[width=\textwidth]{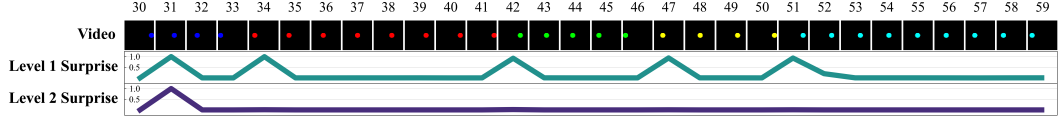}
    \caption{\textbf{Surprise signals on Bouncing Ball}. Uniquely determining future ball colors requires observing the previous six bounces. Once sufficient color history has been observed, the level-2 model is no longer surprised and predicts the future correctly, yielding near-zero surprise. The level-1 model continues to be surprised at each bounce, producing the spikes that define chunk boundaries.}
    \label{fig:high-level-surprise}
\end{figure*}

A natural question is whether deeper hierarchies could yield further gains.
In \proposed{}, an additional level is warranted only if the current top level itself exhibits unpredicted events, since each higher level is intended to model longer-range transitions over the chunks discovered by the level below.
\cref{fig:high-level-surprise} visualizes the level-2 surprise signal across Bouncing Ball trajectories: once the high-level model has observed sufficient color history (six bounces), its surprise stays consistently low and nearly flat, indicating that a third level would have little to model in this setting.
We observe the same trend on the other datasets and therefore restrict \proposed{} to two levels in this work, leaving the study of deeper hierarchies in environments with multi-scale long-range dependencies to future work.

